\documentclass[10pt,twocolumn,letterpaper]{article}

\usepackage[pagenumbers]{wacv} 

\usepackage{graphicx}
\usepackage{amsmath}
\usepackage{amssymb}
\usepackage{booktabs}
\usepackage{microtype}
\usepackage{mathtools}
\usepackage{amsthm}
\usepackage{xargs}     
\usepackage{float}
\usepackage{makecell} 
\usepackage{multirow}
\usepackage[pagebackref,breaklinks,colorlinks]{hyperref}
\usepackage[dvipsnames]{xcolor}
\usepackage[accsupp]{axessibility}

\usepackage[capitalize,noabbrev]{cleveref}
\newcommand{\AlgName}{VioPose}
\newcommand{\DataName}{VioDat}

\definecolor{DarkGreen}{rgb}{0.1,0.6,0.1}
\def\check c{{\color{DarkGreen}\checkmark}}
\def\x x{{\color{red}x}}

\makeatletter
\renewcommand*{\@fnsymbol}[1]{}
\makeatother

\usepackage[numbers]{natbib}

\def\eqref#1{(\ref{eq:#1})}
\def\eqlabel#1{\label{eq:#1}}

\usepackage[capitalize]{cleveref}
\crefname{section}{Sec.}{Secs.}
\Crefname{section}{Section}{Sections}
\Crefname{table}{Table}{Tables}
\crefname{table}{Tab.}{Tabs.}

\def\wacvPaperID{484} 
\def\confName{WACV}
\def\confYear{2025}

\begin{document}

\title{\AlgName: Violin Performance 4D Pose Estimation by\\ Hierarchical Audiovisual Inference}

\author{Seong Jong Yoo$^{*, \dagger}$ \qquad
Snehesh Shrestha$^{*}$\qquad
Irina Muresanu  \qquad
Cornelia Ferm\"{u}ller\\
University of Maryland, College Park \\ 
College Park, MD, 20742, USA \\
\tt\small \{yoosj, snehesh, muresanu, fermulcm\}@umd.edu
\thanks{$^*$ Equal contribution}
\thanks{$^\dagger$ Corresponding author}
} 
%

\maketitle

\begin{abstract}
Musicians delicately control their bodies to generate music. Sometimes, their motions are too subtle to be captured by the human eye. To analyze how they move to produce the music, we need to estimate precise 4D human pose (3D pose over time).
However, current state-of-the-art (SoTA) visual pose estimation algorithms struggle to produce accurate monocular 4D poses because of occlusions, partial views, and human-object interactions. They are limited by the viewing angle, pixel density, and sampling rate of the cameras and fail to estimate fast and subtle movements, such as in the musical effect of vibrato.  We leverage the direct causal relationship between the music produced and the human motions creating them to address these challenges. We propose \AlgName: a novel multimodal network that hierarchically estimates dynamics. 
High-level features are cascaded to low-level features and integrated into Bayesian updates. Our architecture is shown to produce accurate pose sequences, facilitating precise motion analysis, and outperforms SoTA.
As part of this work, we collected the largest and the most diverse calibrated violin-playing dataset, including video, sound, and 3D motion capture poses. Code and dataset can be found in our project page \url{https://sj-yoo.info/viopose/}.
\end{abstract}

\section{Introduction}
\label{sec:intro}
Consider you are at a concert, listening to Dmitri Shostakovich's Waltz No.2 with your eyes closed. Can you visualize the violinist's motion? Now, shift your thoughts to listening to Paganini Caprice no. 24. Would you still envision a similar motion? Likely not. Our brain makes different visual estimates based on auditory input, and vice versa~\cite{mcgurkHearingLipsSeeing1976}. 
Sound provides cues for human motion understanding~\cite{arandjelovicLookListenLearn2017}, especially in the context of music performance~\cite{tsaySightSoundJudgment2013, lerchInterdisciplinaryReviewMusic2020,  essidFusionMultimodalInformation2012,duanAudiovisualAnalysisMusic2019, godoyBodyMovementMusic2009}. 

The interplay between music and human motion has captivated researchers, prompting in the last decade studies along various directions. One direction is generating related music from videos of musical performance, such as piano music from performance videos~\cite{suAudeoAudioGeneration2020} and the sounds of other instruments~\cite{ganFoleyMusicLearning2020a}. Similarly, dance videos can be the basis for music generation~\cite{zhuQuantizedGANComplex2022}, and human motion videos can be used to create rhythmic sounds. Conversely, the transformation from audio to human motion has also been explored, for example, to synthesize novel dance motions from music~\cite{leeDancingMusic2019, marchellusM2CConciseMusic2023, puMusicdrivenDanceRegeneration2022}, or to estimate human skeleton motion from piano~\cite{shlizermanAudioBodyDynamics2018} and violin music~\cite{kaoTemporallyGuidedMusictoBodyMovement2020, shresthaAIMusicGuruMusicAssisted2022}.

\begin{figure}
    \centering
    \includegraphics[width=\linewidth]{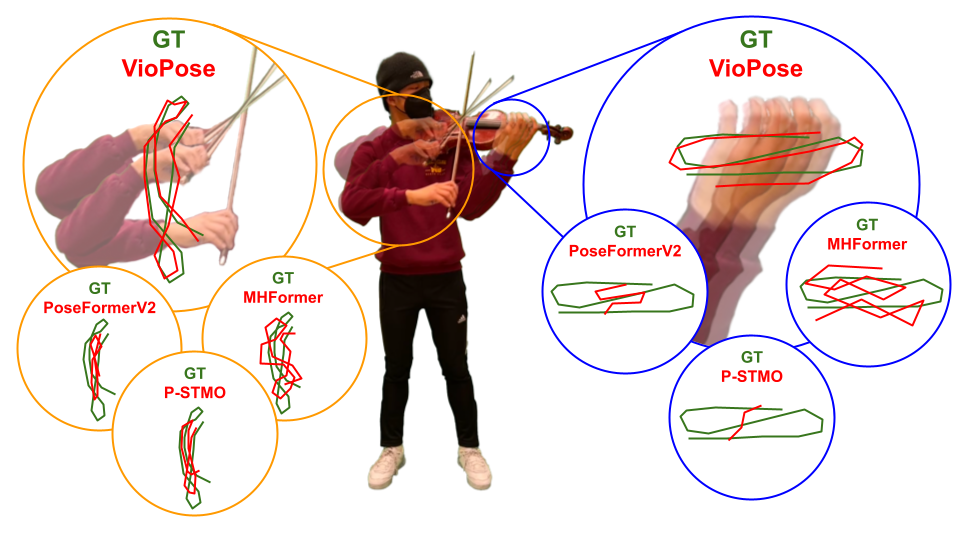}
    \caption{A 4D pose estimation in a violin performance, which features fine grained motion (left hand vibrato, $\approx$ 10 mm perturbation) and large motions (right hand bowing motion). \AlgName \  successfully estimates both  motions, while other approaches fail. See Figs.~\ref{fig:vibrato} and ~\ref{fig:trajectory} for detailed real experimental results.}
    \label{fig:teaser}
    \vspace{-2mm}
\end{figure}
In the literature on human pose estimation, state-of-the-art methods estimate human pose directly from images using different approaches such as RNN architectures~\cite{hossainExploitingTemporalInformation2018, leePropagatingLSTM3D2018}, CNN architectures~\cite{pavlakosOrdinalDepthSupervision2018, pavllo3DHumanPose2019}
attention mechanisms~\cite{liExploitingTemporalContexts2022, liTokenPoseLearningKeypoint2021,zhaoPoseFormerV2ExploringFrequency2023, tang3DHumanPose2023},
and diffusion architectures~\cite{choiDiffuPoseMonocular3D2022, gongDiffPoseMoreReliable2023}. Although these methods, based on local image features, ensure high accuracy in many scenarios, they are often not robust to occlusion. To address this robustness issue, other state-of-the-art methods use physical constraints, e.g. skeleton aware architectures based on  graph neural networks~\cite{ciOptimizingNetworkStructure2019, caiExploitingSpatialTemporalRelationships2019}, algorithms estimating standardized body parameters~\cite{kanazawaEndtoEndRecoveryHuman2018, liHybrIKHybridAnalyticalNeural2021}, or actively change the camera to get better view~\cite{ActPose}.

\begin{figure*}[h]
	\centering
	\includegraphics[width=0.90\linewidth]{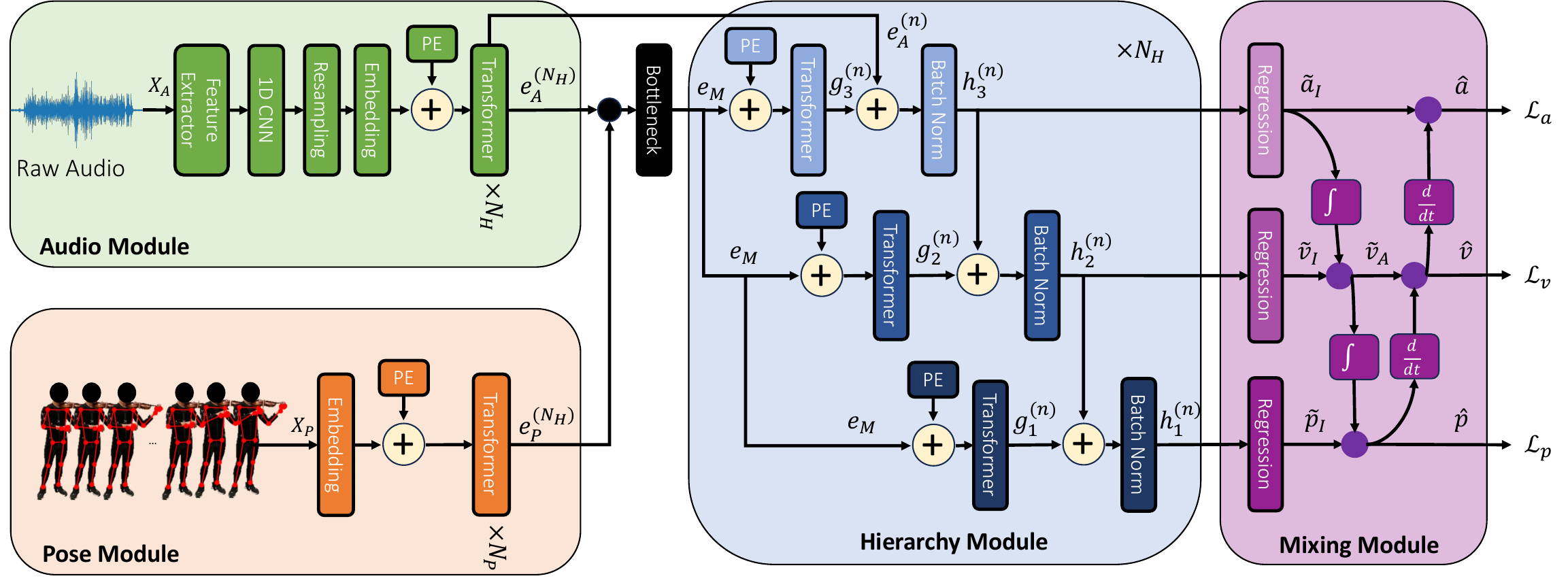}
	\caption{
	    Overview of our proposed \AlgName, a multimodal hierarchical 4D pose estimation pipeline, which receives 2D keypoints computed by an off-the-shelf algorithm and corresponding music-playing audio. Black and purple solid circles represent concatenation and averaging operations, respectively. The architecture consists of three main components: the single modality encoder (green and orange box, $\S$~\ref{sec:single_modality}), the hierarchy module (blue box, $\S$~\ref{sec:hierarchy}), and the mixing module (purple box, $\S$~\ref{sec:mixing}). 
	}
	\label{fig:pipeline}
    \vspace{-2mm}
\end{figure*}
Human pose estimation research has rarely utilized auditory information, relying predominantly on visual data. However, using only visual information to analyze musical performances has limitations. This is because of the challenges in processing music performance videos. First, human movements can be too subtle (e.g., in vibrato, a subtle but quick left-hand motion that fluctuates the pitch slightly to decorate and enhance the musical expression.), which are often ignored because of minor impact on the Mean Per Joint Position Error (MPJPE) (see Fig.~\ref{fig:teaser}). Second, music performance also contains fast and large motion (e.g., bowing motion). Lastly, there are occlusions of the players by their instruments. 
Thus we should leverage sound information to estimate accurate human poses. However, there is a lack of datasets with music and 4D pose. 

To address these challenges, we introduce a new dataset and a novel architecture called \DataName~and  \AlgName. Our multi-modal dataset (Mocap, vision, and audio) has 639 videos of 12 violin players of different gender, ages, size, and skill levels playing a variety of exercises as well as freestyle pieces.
Our algorithm, audiovisual \textbf{\underline{Vio}}lin performance 4D \textbf{\underline{Pose}} estimation, receives audiovisual inputs\textemdash off-the-shelf 2D pose estimates and correlated raw audio\textemdash and estimates 3D pose as output, being trained supervised by ground truth pose collected from motion capture. The hierarchical architecture cascades the high-level information to the lower-level layer and fuses in the way of Bayesian updates. These cross-modal inference modules allow high-quality 3D pose estimation. 

We benchmark \AlgName \ on our dataset \DataName \ against different SoTA networks.   
For a fair comparison, we retrained all other models with \DataName.
Extensive ablation studies were conducted to show the efficacy of proposed modules in the \AlgName. 
In summary, our contributions are as follows:
\begin{itemize}  
    \item We introduce a novel audiovisual monocular 4D pose estimation network that consists of a hierarchical structure. \AlgName \ is able to estimate sophisticated motion including vibrato.
    \item {We introduce the largest fully calibrated and synchronized violin dataset, including three different modalities, diverse participants, playing styles, and levels of expertise.}
    \item {We demonstrate the usefulness of the proposed algorithm through a violin performance analysis task.}
    \AlgName \ successfully analyzes violin performance comparable to ground truth motion results.
\end{itemize}

\section{Related Works}
\label{sec:related}
\paragraph{3D Human pose estimation}
Recent state-of-the-art methods for 3D human pose estimation can be categorized into direct regression and physically constrained methods. Direct regression methods rely on image features to directly locate joint positions~\cite{weiConvolutionalPoseMachines2016, pavlakosOrdinalDepthSupervision2018}. Recent literature uses well-trained 2D keypoint estimation architectures as input and uplifts them to 3D. For instance, ~\citet{pavllo3DHumanPose2019} propose a CNN architecture that exploits the temporal relationship of the input 2D keypoints. Similarly, the Transformer architecture has been used to extract spatial relationships between joints and temporal consistency to uplift from 2D to 3D~\cite{zheng3DHumanPose2021,yangUshapedSpatialTemporal2022, zhaoPoseFormerV2ExploringFrequency2023, tang3DHumanPose2023}. Although these approaches promise high accuracy, they often suffer from occlusion and estimate implausible human poses. Methods that explicitly consider physical constraints in the architectures have used standardized human body information~\cite{kanazawaEndtoEndRecoveryHuman2018}, kinematics~\cite{chenAnatomyAware3DHuman2022, xuDeepKinematicsAnalysis2020}, and physics-based optimization methods~\cite{shimadaPhysCapPhysicallyPlausible2020, hassanResolving3DHuman2019}. \citet{liHybrIKHybridAnalyticalNeural2021} propose a hybrid inverse kinematic architecture that takes advantage of both approaches and ~\cite{liNIKINeuralInverse2023} use an invertible network to combine both. 

However, most methods use visual information only. 
As discussed in the introduction (see Section~\ref{sec:intro}), 
there is a need to utilize audio information in the domain of music performance analysis.

\vspace{-2mm}
\paragraph{Audiovisual learning}
Not only in music, sounds exhibit a strong correlation with motion, paving the way for the design of cross-modal inference architectures. For example, \citet{suPhysicsDrivenDiffusionModels2023} introduces a novel physics-driven diffusion model to estimate impact sounds from corresponding videos. \citet{liLearningVisualStyles2022} employ an audiovisual GAN framework to generate images of scenes, like rain or snowfall, that are in sync with specific sounds. Similarly, generating a talking face from speech, utilizing the combination of audio and visual features has been highlighted~\cite{SynthesizingObamaLearning, parkSyncTalkFaceTalkingFace2022, zhouPoseControllableTalkingFace2021, LipSyncExpert}. Moreover, \citet{ginosarLearningIndividualStyles2019} addresses the speech-to-gesture generation problem using a GAN model. 
However, since different from music performance, speech and gesture are correlated but not {\it causally related}, cross-modality may not be justified.
Extending beyond these applications, multimodal audiovisual architectures have been developed to tackle complex challenges such as sound source separation in the cocktail party problem~\cite{zhaoSoundMotions2019, zhaoSoundPixels2018, afourasSelfsupervisedLearningAudioVisual2020, liAudioVisualEndtoEndMultiChannel2023}, action recognition~\cite{chenMMViTMultiModalVideo2022, rajasekarJointCrossAttentionModel2022, kazakosEPICFusionAudioVisualTemporal2019}, navigation~\cite{chenSoundSpacesAudioVisualNavigation2020a, ganLookListenAct2020}, and depth estimation~\cite{gaoVisualEchoesSpatialImage2020}.

Most audiovisual fusion methods utilize either early or mid-feature fusion architectures. Our architecture adopts a mid-fusion strategy, but unlike other architectures, auditory information is computed as a prior for high-level posterior computation.

\begin{table*}[ht]
    \centering
    \resizebox{2\columnwidth}{!}{
    \begin{tabular}{l|ccccc|ccc|ccc|ccc}
        \Xhline{1pt}
        \multicolumn{1}{l|}{} & \multicolumn{5}{c|}{\bf{General Info}} & \multicolumn{3}{c|}{\bf{Pose Info}} & \multicolumn{3}{c|}{\bf{Video Info}} & \multicolumn{3}{c}{\bf{Audio Info}}\\
        \hline
        \bf{Dataset}    & \bf{Type}     & \bf{Interact}     & \bf{Fine}     & \bf{\# Subj}     & \bf{Div}      & \bf{3D}      & \bf{2D}     & \bf{Calib} & \bf{\# Vids}  & \bf{Len (m)}  & \bf{\# Cams}      & \bf{\# Mics}     & \bf{Audio}    & \bf{Causal}   \\
        \hline
        
        HumanEva \cite{Sigal:IJCV:10b}   & M & \x x{}   & \x x{}    & 4     & \x x{}     & \check c{}    & \check c{}    & \check c{}    & 168   & 22    & 4    & 0     & \x x{}    & \x x{} \\
        
        Human3.6M \cite{6682899}         & M & \x x{}   & \x x{}    & 11    & \check c{}     & \check c{}    & \check c{}    & \check c{}    & 660   & 298    & 4    & 0     & \x x{}    & \x x{} \\
        
        \hline
        
        GrooveNet \cite{alemi2017groovenet} & D     & \x x{}   & \x x{}  & 11   & \x x{}   & \check c{}    & \x x{}    & \x x{}    & 0    & 0    & 0    & 1 & \check c{}   & \x x{} \\ 
        
        DanceNet  \cite{zhuang2020music2dance} & D  & \x x{}   & \x x{}  & 2    & \x x{}   & \check c{}    & \x x{}    & \x x{}    & 0    & 0    & 0    & 1 & \check c{}    & \x x{} \\
        
        AIST++ \cite{liAIChoreographerMusic2021}              & D  & \x x{}    & \x x{}    & 30   & \x x{}   & \check c{}    & \check c{}    & \check c{} & 1408  & 312    & 9    & 1   & \check c{}  & \x x{} \\

        \hline
        
        QUARTET \cite{Papiotis:PhdThesis2016}    & P & \check c{} & \check c{} & 4  & \x x{}   & \x x{}    & \x x{}    & \x x{}    & 30     & 29    & 1     & 6     & \check c{}    & \check c{} \\
        
        TELMI \cite{volpe2017multimodal}         & P & \check c{} & \check c{}  & 4 & \x x{}   & \x x{}    & \x x{}    & \x x{}    & 292    & 105   & 2     & 3     & \check c{}    & \check c{} \\ 

        URMP    \cite{li2018creating}       & P  & \check c{} & \check c{}   & 22   & \x x{}   & \x x{}    & \x x{}    & \x x{}    & 44     & 78    & 1     & 1     & \check c{}    & \check c{} \\
        
        \hline 
        
        \bf{\DataName \ (Ours)}       & \bf{P} & \check c{}  & \check c{}  & \bf{12} & \check c{}    & \check c{}    & \check c{}    & \check c{}    & \bf{639}   & \bf{133}  & \bf{4}   & \bf{4}    & \check c{}  & \check c{} \\
        
        \Xhline{1pt}
    \end{tabular} 
    }
    \caption{
    Comparison of 3D human kinematic pose and audio datasets. Symbols (\check c{}) means fully satisfies, (\x x{})  does not satisfy, and ($\sim$) partially satisfies the field topic. Dataset \underline{\textit{Type}} include General motion (\underline{\textit{M}}),  Dance (\underline{\textit{D}}), and Playing (\underline{\textit{P}}) instruments. \underline{\textit{Interact}} refers to people interacting with objects, thus making the dataset difficult for pose estimation. \underline{\textit{Fine}} refers to the motion velocity and complexity, where slow-moving and large change is easier to observe. In contrast, fast motion and small movements are much more challenging due to motion blur and pixel saturation. In Dance datasets (\underline{\textit{D}}) the audio is not produced by the motions and therefore does not have a \underline{\textit{\textbf{Causal}}} relationship. The Musical instrument playing (\underline{\textit{P}}) datasets QUARTET and TELMI contain raw motion capture markers, audio (\underline{\textit{\# Mics}}), and video (\underline{\textit{\# Vids}}), and URMP has audio and video. These datasets lack accurate 3D human kinematic and joint information and therefore \underline{\textit{3D}} and \underline{\textit{2D}} Pose Ground Truth, and they do not have the information on the video calibration (\underline{\textit{Calib}}) making it difficult to use and compare with Pose from video. Multiple subjects (\underline{\textit{\# Subj)}} of diverse (\underline{\textit{Div}}) races, genders, ages, height, body type, and skill levels are important for a reliable dataset. \DataName is the only dataset that has a fully calibrated full-body 3D and 2D human pose with synchronized audio recorded from diverse subjects from multiple viewpoints.}
    \label{tab:dataset_comparison}
    \vspace{-2mm}
\end{table*}

\paragraph{Audiovisual music performance video analysis}
While sound plays a key role, visual cues also contain abundant information for understanding music performance~\cite{essidFusionMultimodalInformation2012, tsaySightSoundJudgment2013}. However, analyzing music performance videos is challenging due to the subtle or rapid movements of body parts, important limbs being highly occluded or truncated by musical instruments or other people, and the involvement of multiple musicians in a single frame~\cite{duanAudiovisualAnalysisMusic2019}. To address these challenges, several audiovisual architectures have been proposed. For example, \citet{zhangAutomaticMusicTranscription2009} propose an audiovisual fusion algorithm for violin onset detection that combines feature-level early fusion with decision-level late fusion. \citet{okaMarkerlessPianoFingering2013} present a method for piano fingering detection using depth cameras and MIDI signals. Beyond audiovisual learning, \citet{gaoAutomatedViolinBowing2023a} address the problem of violin bowing technique classification using an FMCW radar system. However, the absence of high-quality audiovisual datasets hinders the development of fundamental architectures.

It is important to note that estimating accurate 3D pose is fundamental to music performance analysis. Based on joint trajectories, bowing technique classification can be accomplished~\cite{dalmazzoMachineLearningApproach2018, dalmazzoBowingGesturesClassification2019}, or instant feedback for beginners can be provided~\cite{percivalEffectiveUseMultimedia2007, blancoRealTimeSoundMotion2021}. 

\section{Method}
\label{sec:method}
\AlgName \ learns a hierarchy of motion dynamic embeddings from audiovisual features. The network receives a sequence of 2D pose estimates from an off-the-shelf algorithm (i.e. MediaPipe~\cite{lugaresiMediaPipeFrameworkBuilding2019}) and corresponding raw audio as input, and fuses them to estimate 3D pose sequences, velocity and acceleration as the final output of architecture. 
Firstly, the network extracts audio latent vectors from audio features using 1D CNN and Transformer layers, and temporal 2D human body structure consistency latent vectors by Transformer layers ($\S$ \ref{sec:single_modality}), respectively. The architecture then combines computed audiovisual latent vectors through a bottleneck layer. 
Secondly, the combined audiovisual latent vectors are fed into the hierarchy module to learn human motion dynamics. Specifically, the high-level features are cascaded to the lower layer and combined in a way similar to Bayesian updates (e.g., $\log p(x|y) \approx \log p(y|x) + \log p(x)$). This method efficiently combines higher-level motion dynamics features with lower-level motion dynamics features ($\S$ \ref{sec:hierarchy}). 
Finally, the estimated motion dynamics are integrated by bidirectional averaging ($\S$ \ref{sec:mixing}).  Figure \ref{fig:pipeline} shows an overview of our proposed pipeline.

Note that in our application, the audio is directly generated by human motion. This strong causal relationship provides significant cues for estimating 4D pose, especially subtle motion, such as vibrato (see Fig.~\ref{fig:vibrato}). 

\subsection{Problem Statement}
Using any off-the-shelf 2D pose estimation pipeline (i.e. MediaPipe~\cite{lugaresiMediaPipeFrameworkBuilding2019}), we extract noisy 2D poses $X_{M}\in \mathbb{R}^{f\times J \times 2}$, where $J$ is the number of joints and $f$ is the initial input video frames. In addition to the 2D pose, we utilize correlated audio information $X_{A} \in \mathbb{R}^{F}$, where $F \gg f$. The goal of the task is to generate accurate 3D human poses $X_{P} \in \mathbb{R}^{f\times J\times 3}$.

\subsection{Single Modal Feature Embedding}
\label{sec:single_modality}
\paragraph{Body consistency feature embedding} We compute 2D keypoints from MediaPipe~\cite{lugaresiMediaPipeFrameworkBuilding2019}. We then tokenize the input 2d joints ($X_{P}\in \mathbb{R}^{B \times f \times D}$) through a learnable linear embedding layer followed by a learnable positional embedding, where $B$ is the batch size, $f$ is the number of input frames, and $D$ is the dimension of the latent vectors. Multiple transformer layers compute temporal 2D body consistency latent vector ($e_{P}^{(n)}\in \mathbb{R}^{B\times f \times D}, n \in [N_P]$) as follows: 

\begin{equation}
\eqlabel{eq:transformer}
\begin{split}
    \hat{e}_{P}^{(n-1)}&=e_{P}^{(n-1)}+\mathrm{MHA}(\mathrm{LN}(e_{P}^{(n-1)}))  \\ 
    e_{P}^{(n)} &= \hat{e}_{P}^{(n-1)} + \mathrm{FFN}(\mathrm{LN}(\hat{e}_{P}^{(n-1)})) 
\end{split}
\end{equation}
where $\mathrm{MHA}$, $\mathrm{LN}$, and $\mathrm{FFN}$ denote \textit{multi-head attention layer}, \textit{layer normalization}, and \textit{fully connected layer}, respectively. $N_P$ is the number of Transformer blocks. This way, the model can learn the temporal consistency of 2D human body information~\cite{zheng3DHumanPose2021}. 

\begin{figure}[t]
	\centering
	\includegraphics[width=\linewidth]{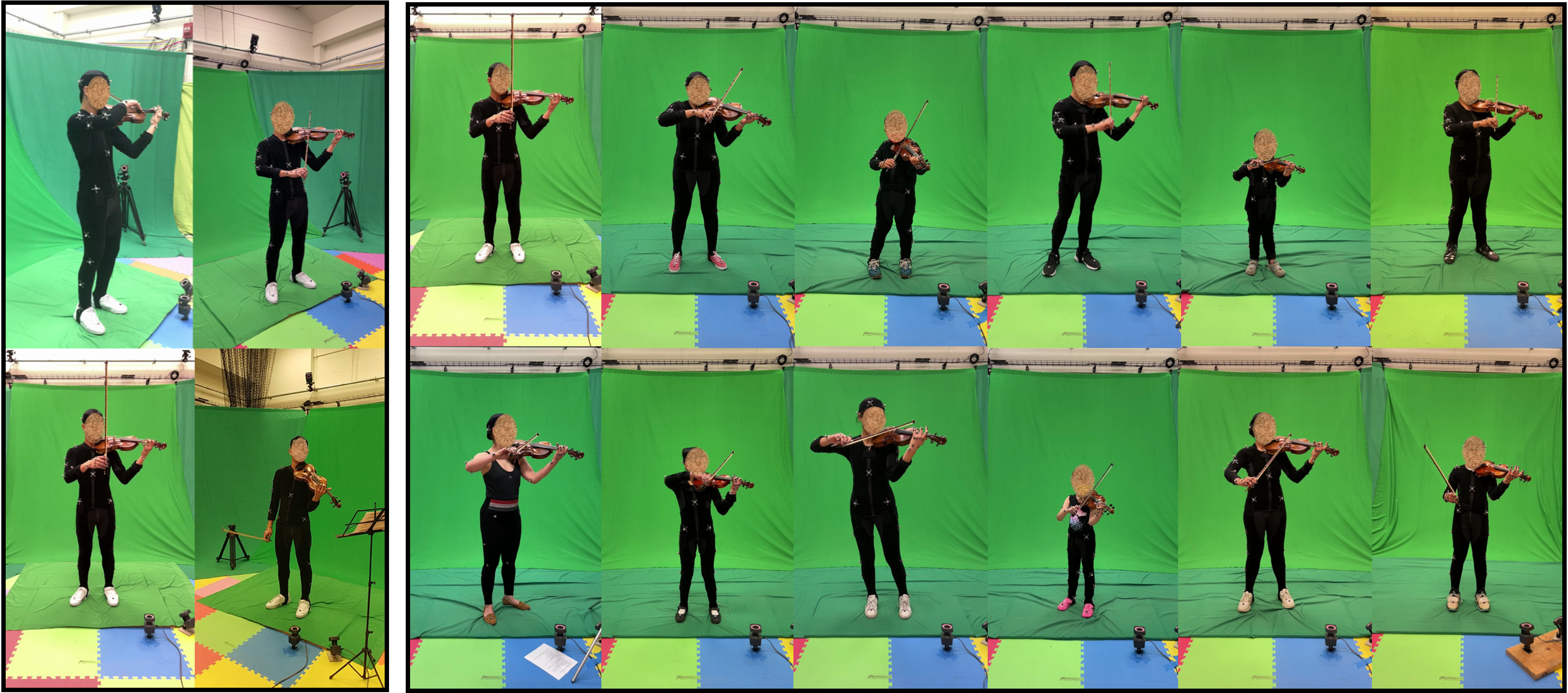}
	\caption{
 Our dataset was recorded from 4 different camera views (left figure) with video at $\approx$ 30 FPS and synchronized audio using smartphones. We have a total of 12 people with different gender, age, height, violin size, and body type (right figure).
	}
	\label{fig:dataset}
\vspace{-2mm}
\end{figure}
\paragraph{Audio feature embedding} 
From the raw audio, we first compute a 35-dim audio feature vector consisting of the 1-dim envelope, 20-dim MFCC, 12-dim Chroma, 1-dim one-hot peaks, and 1-dim RMS~\cite{liAIChoreographerMusic2021, marchellusM2CConciseMusic2023}. We use a 1D CNN to encode audio features and resample $F$ to $f$ (Table. ~\ref{tab:abl_audio} shows the effects of different audio sampling rates and inputs). We then extract the audio latent vector $e_{A}^{(n)}\in \mathbb{R}^{B\times f\times D}$ using Eq.\eqref{eq:transformer}. Each latent vector ($e_{A}^{(n)}$) feeds into the highest layer of the hierarchy module. Naturally, audio contains information on human dynamics because it is generated by the dynamic interaction between humans and instruments. Therefore, we treat audio latent vectors as prior information to compute the human motion acceleration ($\S$ ~\ref{sec:hierarchy}). 

Finally, the audiovisual feature, $e_{M}\in \mathbb{R}^{B\times f \times D}$, is computed by concatenation of the body structure latent vectors and audio latent vectors, followed by a bottleneck layer (block box in Fig~\ref{fig:pipeline}).

\subsection{Hierarchy Module}
\label{sec:hierarchy}
To estimate precise 4D pose while considering both fine (see Fig. ~\ref{fig:vibrato}) and large motions (see Fig. ~\ref{fig:trajectory}) simultaneously, we devise a hierarchical architecture as shown in the blue box in Fig.~\ref{fig:pipeline}. The hierarchy structure consists of three layers, each directly mapped to estimate motion dynamics: the highest layer to acceleration, the middle layer to velocity and the lowest layer to pose. 
Higher dynamics information works as prior information of lower dynamics computation. We implicitly design the log Bayesian update rules as a cascade summation structure. 
Each layer receives information from the upper layer and combines it with the Transformer output. 
Formally, we design a cascade updating structure as follows:
\begin{align}
    g_{i}^{(n)} &= \text{Transformer}(h_{i}^{(n-1)}) \eqlabel{eq:hierarchy}  \\ 
    h_{i}^{(n)} &= \text{Norm}(g_{i}^{(n)} + h_{i+1}^{(n)}) \eqlabel{eq:sum},
\end{align}
where $i\in [3], n \in [N_{H}], h_{i}^{0}=e_{M}$,  $h_{4}^{(n)}=e_{A}^{(n)}$, $\text{Norm}(x)$ is a batch norm, and  $\text{Transformer}(x)$ is using self-attention described in Eq.~\eqref{eq:transformer}. Note that $N_H$ is the number of Transformer blocks used in audio feature embedding module.

\subsection{Bidirectional Mixing Module}
\label{sec:mixing}
In this module, we bidirectionally mix the estimated motion dynamics as shown in the purple box in Fig.~\ref{fig:pipeline}.   Firstly, the architecture generates an initial estimated acceleration ($\tilde{a}_I)$) from the hierarchy module and predicts velocity by integrating it. Then, this estimate is combined with the initial estimated velocity ($\tilde{v}_I$) from the hierarchy module, and a similar computation is done to pose estimation. Secondly, we start by differentiating the estimated pose to compute the predicted velocity and then compute the final value with the previous velocity. A similar computation is done for the acceleration as well. Formally, the bidirectional mixing module is as follows:
\begin{align}
        \tilde{v}_{A} &= \frac{1}{2} (\tilde{v}_{I} + \int \tilde{a}_I dt) \eqlabel{eq:vel_int}\\ 
        \hat{p} &= \frac{1}{2} (\tilde{p}_{I} + \int \tilde{v}_A dt) \eqlabel{eq:pose_int}\\ 
        \hat{v} &= \frac{1}{2} (\tilde{v}_{A} + d \hat{p}/dt) \eqlabel{eq:vel_diff}\\
        \hat{a} &= \frac{1}{2} (\tilde{a}_{I} + d \hat{v}/dt) \eqlabel{eq:acc_diff}
\end{align}
Note that the order of computation is Eq~\eqref{eq:vel_int} to Eq~\eqref{eq:acc_diff}.

By introducing the bidirectional mixing module, we explicitly combine estimated motion dynamics. At inference time, we further combine the final estimated motion dynamics using a Kalman filter. In other works, architectures also learn velocity, but the estimation is done at the middle layers~\cite{einfaltUpliftUpsampleEfficient2023, zhangMixSTESeq2seqMixed2022}. However, these approaches often fail to generate fine and fast motions precisely. We empirically evaluate the efficacy of the bidirectional module in Table ~\ref{tab:abl}.



\subsection{Loss Function}
Our loss function consists of three losses: the final pose estimation loss $\mathcal{L}_{p}$, the velocity loss $\mathcal{L}_{v}$, and the acceleration loss $\mathcal{L}_{a}$.  The pose estimation loss $\mathcal{L}_{p}$ is defined using the MPJPE (Mean Per Joint Position Error) and the velocity and acceleration losses are defined using the max-cosine similarity~\cite{hafnerDeepHierarchicalPlanning2022a}, given as follows:
\begin{equation}
\eqlabel{eq:loss}
\begin{split}
    \mathcal{L}_{p}(\hat{p}, p^{GT}) &= \frac{1}{f} \sum_{i=1}^{f}(\sum_{j=1}^{J} || \hat{p}_{i, j}-p_{i, j}^{GT}||_{2}) \\ 
    \mathcal{L}_{v}(\hat{v}, v^{GT}) &= \frac{1}{f} \sum_{i=1}^{f}(\sum_{j=1}^{J} (\hat{v}/V)^{\top}(v^{GT}/V) \\ 
    \mathcal{L}_{a}(\hat{a}, a^{GT}) &= \frac{1}{f} \sum_{i=1}^{f}(\sum_{j=1}^{J} (\hat{a}/A)^{\top}(a^{GT}/A)\\
    \mathcal{L} &= \mathcal{L}_{p} + \lambda_v \mathcal{L}_{v} + \lambda_a \mathcal{L}_{a},
\end{split}
\end{equation}
where $V=\max (\|\hat{v}\|, \| v^{GT}\|), A=\max (\|\hat{a}\|, \| a^{GT}\|), \lambda_{v}$ and $ \lambda_{a}$ are balancing weights. The max-cosine similarity considers both direction and magnitude, which tends to work better than the L2 loss.

\begin{figure}[t]
	\centering
	\includegraphics[width=1.0\linewidth]{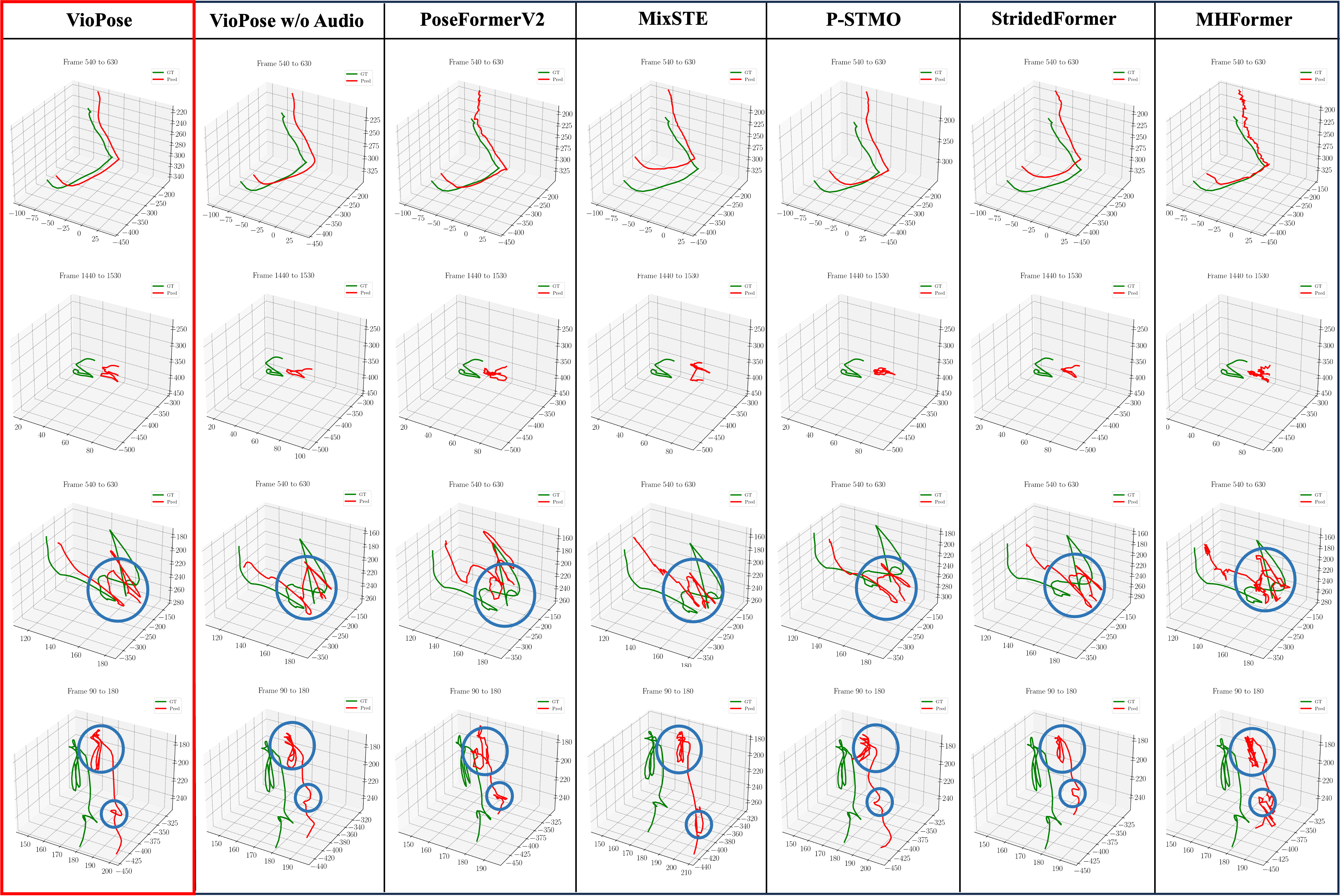}
	\caption{
	    Predicted right wrist trajectories (red line) and the ground truth 3D trajectories (green line) after Procrustes projection for better comparison. Each graph contains 90 frames (3 seconds).
	}
	\label{fig:trajectory}
    \vspace{-2mm}
\end{figure}

\begin{table}[b]
\vspace{-2mm}
    \centering
    \resizebox{\columnwidth}{!}{
    \begin{tabular}{l| c| c| c | c| c }
     \Xhline{1pt}
     \textbf{Method} & \textbf{MPJPE $\downarrow$} & \textbf{P-MPJPE $\downarrow$} & \textbf{MPJVE $\downarrow$} & \textbf{MPJAE $\downarrow$} & \textbf{DTW $\downarrow$} \\
    \hline
    
    PoseFormerV2 (T=81)~\cite{zhaoPoseFormerV2ExploringFrequency2023} 
    & 54.87 & 33.51 & 3.16 & 3.10 & 7.57
    \\
    
    MixSTE (T=243) ~\cite{zhangMixSTESeq2seqMixed2022}
    & 49.92 & 33.71 & 2.60 & 3.04 & 7.26
    \\
    
    P-STMO (T=243) ~\cite{shanPSTMOPretrainedSpatial2022} 
    & 46.09 & 28.42 & 1.96 & 1.93 & 6.90
    \\
    
    StridedFormer (T=351) ~\cite{liExploitingTemporalContexts2022}
    & 49.12 & 29.59 & 1.90 & 1.69 & 7.08
    \\
    
    MHFormer (T=351) ~\cite{liMHFormerMultiHypothesisTransformer2022} 
    & 49.44 & \textcolor{blue}{27.78} & 5.90 & 9.41 & 7.13
    \\
    \hline
    \AlgName \ w/o audio (T=150) 
    & \textcolor{blue}{44.78} & 28.66 & \textcolor{blue}{1.78} & \textcolor{blue}{1.25} & \textcolor{blue}{6.80}
    \\
    \AlgName \  (T=150)
    & \textcolor{red}{43.60} & \textcolor{red}{27.77} & \textcolor{red}{1.57} & \textcolor{red}{1.02} & \textcolor{red}{6.67} 
    \\
    
    \Xhline{1pt}
    \end{tabular}
    }
    \caption{Comparison of MPJPE, MPJVE, MPJAE, P-MPJPE, and DTW (dynamic time warping) with other SoTA methods on the proposed data set, where $T$ is the length of the input frame. \textcolor{red}{Red} represents the best value and \textcolor{blue}{blue} is the second best.}
    \label{tab:result_human3d_pose}
\end{table}

\begin{table*}[h!]
    \centering
    \resizebox{2\columnwidth}{!}{
    \begin{tabular}{l|c c c | c c c|c c c| c c c | c c c | c c c  }
         \Xhline{1pt}
            & \multicolumn{9}{c|}{\textbf{Left Body}} & \multicolumn{9}{c}{\textbf{Right Body}}\\
         \cline{2-19}
            
            & \multicolumn{3}{c|}{\textbf{Upper Limb}}  & \multicolumn{3}{c|}{\textbf{Hand}} & \multicolumn{3}{c|}{\textbf{Lower Limb}} & \multicolumn{3}{c|}{\textbf{Upper Limb}} & \multicolumn{3}{c|}{\textbf{Hand}} &\multicolumn{3}{c}{\textbf{Lower Limb}} \\
        \cline{2-19}
        \textbf{Method}
            & \textbf{Sh.} & \textbf{El}. & \textbf{Wr}. & \textbf{In.} & \textbf{Pi.} &\textbf{Th.} 
            & \textbf{Pe.} &\textbf{Kn.} & \textbf{An.}  
            & \textbf{Sh.} &\textbf{El.} & \textbf{Wr.} & \textbf{In.} & \textbf{Pi.} &\textbf{Th.} 
            & \textbf{Pe.} &\textbf{Kn.} & \textbf{An.} \\
            
        \Xhline{1pt}
            PoseFormerV2 (T=81)~\cite{zhaoPoseFormerV2ExploringFrequency2023}
            &46.08 & 48.38  &66.98  &71.73 & 74.25 & 68.06
            & 13.35  & 47.14 & 64.15
            & 50.43 & 50.43  & 63.14 & 66.64 & 67.97 & 69.03
            & 13.35  & 46.92 & 56.61
            \\
            
            MixSTE (T=243) ~\cite{zhangMixSTESeq2seqMixed2022}
            & 33.91 & 50.81 & 68.20 & 66.81 & 68.37 & 63.13
            & 11.74 & 46.80 & \textcolor{red}{50.63}
            & 43.09 & 51.30 & 54.50 & 58.91 & 59.83 & 62.97
            & 11.74 & 48.87 & 46.93 
            \\ 
            
            P-STMO (T=243) ~\cite{shanPSTMOPretrainedSpatial2022}
            & 33.80 & 46.40 & 63.29 & 64.87 & 67.38 & 61.84
            & 10.40 & 44.39 & 51.13
            & \textcolor{blue}{37.63} & \textcolor{blue}{45.26} & 48.44 & 51.37 & 52.43 & 52.38
            & 10.40 & 42.98 & \textcolor{blue}{45.21}
            \\
            
            StridedFormer (T=351) ~\cite{liExploitingTemporalContexts2022} 
            & 37.60 & \textcolor{blue}{46.01} & 65.52 & 66.75 & 67.93 & 65.45
            & 10.30 & 45.44 & 56.14
            & 43.13 & 50.44 & 50.44 & 53.69 & 54.61 & 55.65
            & 10.30 & 50.26 & 54.74
            \\ 
            
            MHFormer (T=351) ~\cite{liMHFormerMultiHypothesisTransformer2022} 
            & 38.56 & 50.44 & 64.98 & 66.08 & 68.26 & 64.27
            & 10.91 & 43.84 & 60.58
            & 43.68 & 47.11 & 53.74 & 55.18 & 55.80 & 57.12
            & 10.91 & 46.61 & 51.86
            \\
            \hline
            \AlgName \ w/o Audio (T=150)
            & \textcolor{blue}{32.00} & 47.01  & \textcolor{blue}{59.96} & \textcolor{blue}{60.68} & \textcolor{blue}{63.27} & \textcolor{blue}{58.47}
            &\textcolor{blue}{10.10} & \textcolor{red}{40.32} & \textcolor{blue}{50.82}
            & 39.12 & 46.60 & \textcolor{blue}{46.96}  & \textcolor{blue}{49.58} & \textcolor{blue}{50.50} & \textcolor{blue}{52.24}
            & \textcolor{blue}{10.10} & \textcolor{red}{41.63} & 46.62
            \\
            
            \AlgName \  (T=150)
            & \textcolor{red}{30.10} & \textcolor{red}{45.30} & \textcolor{red}{59.75} & \textcolor{red}{59.80} & \textcolor{red}{62.52} & \textcolor{red}{57.90}
            & \textcolor{red}{9.99} & \textcolor{blue}{41.58} & 51.52
            & \textcolor{red}{36.34} & \textcolor{blue}{46.42} & \textcolor{red}{44.29} & \textcolor{red}{47.39} & \textcolor{red}{48.06} & \textcolor{red}{48.57}
            & \textcolor{red}{9.99} & \textcolor{blue}{42.22} & \textcolor{red}{42.97}
            \\
        \Xhline{1pt}
    \end{tabular}
    }
    \caption{Comparison of the 3D position error of each joint: Shoulder (\underline{\it{Sh.}}), Elbow (\underline{\it{El.}}), Wrist (\underline{\it{Wr.}}), Index (\underline{\it{In.}}), Pinky (\underline{\it{Pi.}}), Thumb (\underline{\it{Th.}}), Pelvis (\underline{\it{Pe.}}), Knee (\underline{\it{Kn.}}) and Ankle (\underline{\it{An.}}) with other SoTA methods. \textcolor{red}{Red} represents best value and \textcolor{blue}{blue} is second best.
    }
    \label{tab:result_human3d_pose_joints}
\vspace{-2mm}
\end{table*}

\section{Experiments}
\label{sec:exp}
\subsection{Implementation Details}
We employed MediaPipe~\cite{lugaresiMediaPipeFrameworkBuilding2019} to extract the initial noisy 2D keypoints from the input video. The number of input frames ($f$) is 150, and the number of audio frames ($F$) is 500. We cropped the video and audio into windows of 3 seconds and 1-second hop resulting in 24,758 samples. The 1D CNN layers of the audio module have filter size $\{64, 32, 16\}$ and kernel size $\{3, 3, 3\}$ with strides $\{1, 1, 1\}$. We used three layers in the Transformer and three stacked hierarchical modules ($N_P=3$ and $N_H=3$). We used 256 hidden dimensions for the Transformer. We trained the network with the Adam optimizer for 150 epochs with a mini-batch size of 64. The learning rate is set to $1\times 10^3$ and reduced to $5\times 10^4$ and $1\times 10^4$ at  50 and 100 epochs, respectively. We divided the dataset into training, validation, and test sets by participants. The test set contains 3 participants (an advanced male adult, a novice female teenager, and a novice male child) and has $\approx$ 30 \% of the total dataset. We randomly mixed the other 70 \% data and divided it into 90 \% for the training and 10 \% for the validation.

\begin{figure}[t]
	\centering
	\includegraphics[width=1.0\linewidth]{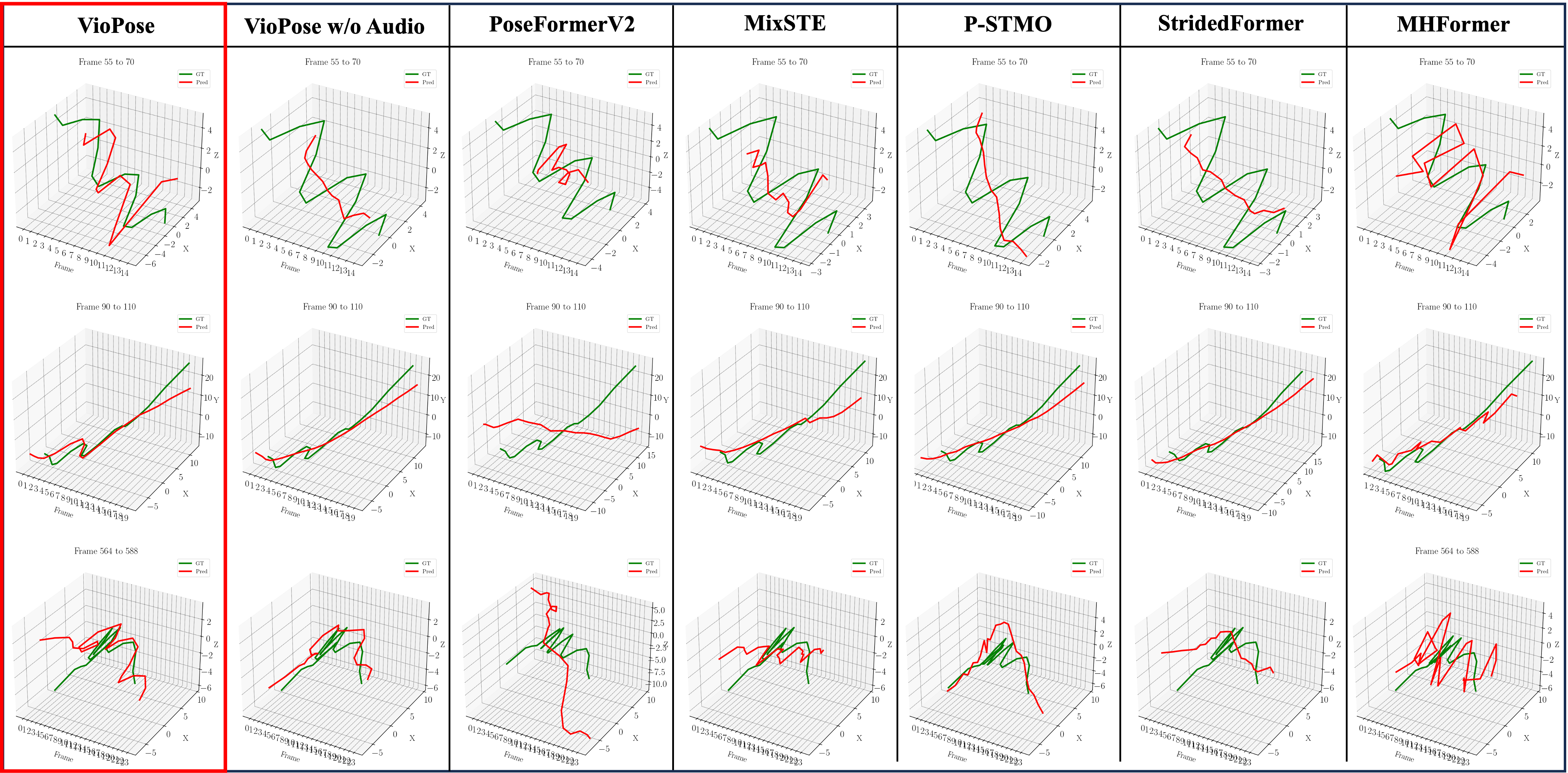}
	\caption{
	    Predicted left hand trajectories (red line) and the ground truth 3D trajectories (green line) in the vibrato movement after center alignment for better comparison. Most of the SoTAs estimate simple straight lines, but \AlgName \ is able to estimate the fine vibrato motion. Note that MHFormer looks like it is able to estimate vibrato but the movement contains high jitter estimation. We can verify this from the trajectories in Fig.~\ref{fig:trajectory}, or the MPJVE and MPJAE metrics in Table~\ref{tab:result_human3d_pose}). 
	}
	\label{fig:vibrato}
    \vspace{-2mm}
\end{figure}

\subsection{Dataset}
We collected a new violin music dataset called \DataName\ that can be used for accurate pose estimation of players. It is the largest multimodal dataset (MoCap data, 4 cameras, and 4 audio sensors) and includes 12 participants with varying levels of expertise from novice (7 participants) to advanced players (5) playing simple bowing exercises, \'etudes, and general musical pieces with simple to complex bowing techniques totaling 639 videos. In the first part of the data collection, players followed a strict protocol, and in the second half, they played varying styles and types of music. Data were recorded from four 720p resolution camera views as shown in Fig.\ref{fig:dataset} that were fully calibrated and synchronized with the 12-camera Vicon 3d motion capture system (mocap). Video data was sampled at approximately 30 frames per second (fps), audio at 44.1 kHz, and mocap at 100 Hz with 0.1mm accuracy. All participants wore mocap suits and had carefully placed retro-reflective markers on their hands and fingers (although the finger data is not used in this paper) totaling 96 markers, which were post-processed to compute the human body kinematic models by extracting the accurate joint center coordinates. The dataset contains 2D and 3D ground truth and the corresponding audio. To highlight the need for a multimodal, diverse, fully calibrated music dataset, a comparison to generic motion datasets, dance datasets, and music datasets is presented in Table \ref{tab:dataset_comparison}.

\subsection{Experimental Results}
We used the Mean Per Joint Position Error (MPJPE), MPJPE after Procrustes analysis (P-MPJPE), Mean Per Joint Velocity Error (MPJVE), Mean Per Joint Acceleration Error (MPJAE), and Dynamic Time Warping (DTW). MPJVE and MPJAE measure smoothness. DTW and MPJAE measure the similarity between the generated trajectory and the ground truth trajectory. 
We found that subtle movements, such as the vibrato, present a challenge for accurate motion estimation.
Thus, a model that only estimates the average trajectory position may yield a low MPJPE but a high MPJVE, MPJAE, and DTW. Therefore, we need to consider all the metrics to understand the model's fidelity.

We benchmarked our results against other state-of-the-art architectures on \DataName. To ensure a fair comparison, we retrained these architectures using our dataset.  
Table ~\ref{tab:result_human3d_pose} presents the experimental results. We established the hip position (calculated as the average value of the pelvis) as a reference. 
Still, our algorithm, \AlgName, demonstrated superior performance; compared to the best values of others, we achieved a 5.40 \% improvement in MPJPE. Remarkably, \AlgName \ achieved a significant 17.37 \% and 39.64 \% improvement in MPJVE and MPJAE values, respectively. Furthermore, \AlgName \ without the audio model also achieved the second best performance in all the metrics except P-MPJPE. 

Unlike other datasets, \DataName \ contains extreme motion disparity; the lower body limbs mostly stay still but the upper body limbs dynamically move. Because of these special conditions, it is preferable to evaluate joint-wise position accuracy rather than MPJPE, see Table~\ref{tab:result_human3d_pose_joints}. Based on the result, we observed that the most accuracy gains came from estimating accurate upper body limbs. For example, we achieved a maximum 7.82 \% accuracy gain in left index estimation and 8.57 \% accuracy gain in right wrist estimation compared to other best SoTA results. These results demonstrate  that our proposed architecture successfully estimates the complex motion dynamics of violin playing. 

\begin{figure}[t]
	\centering
	\includegraphics[width=\linewidth]{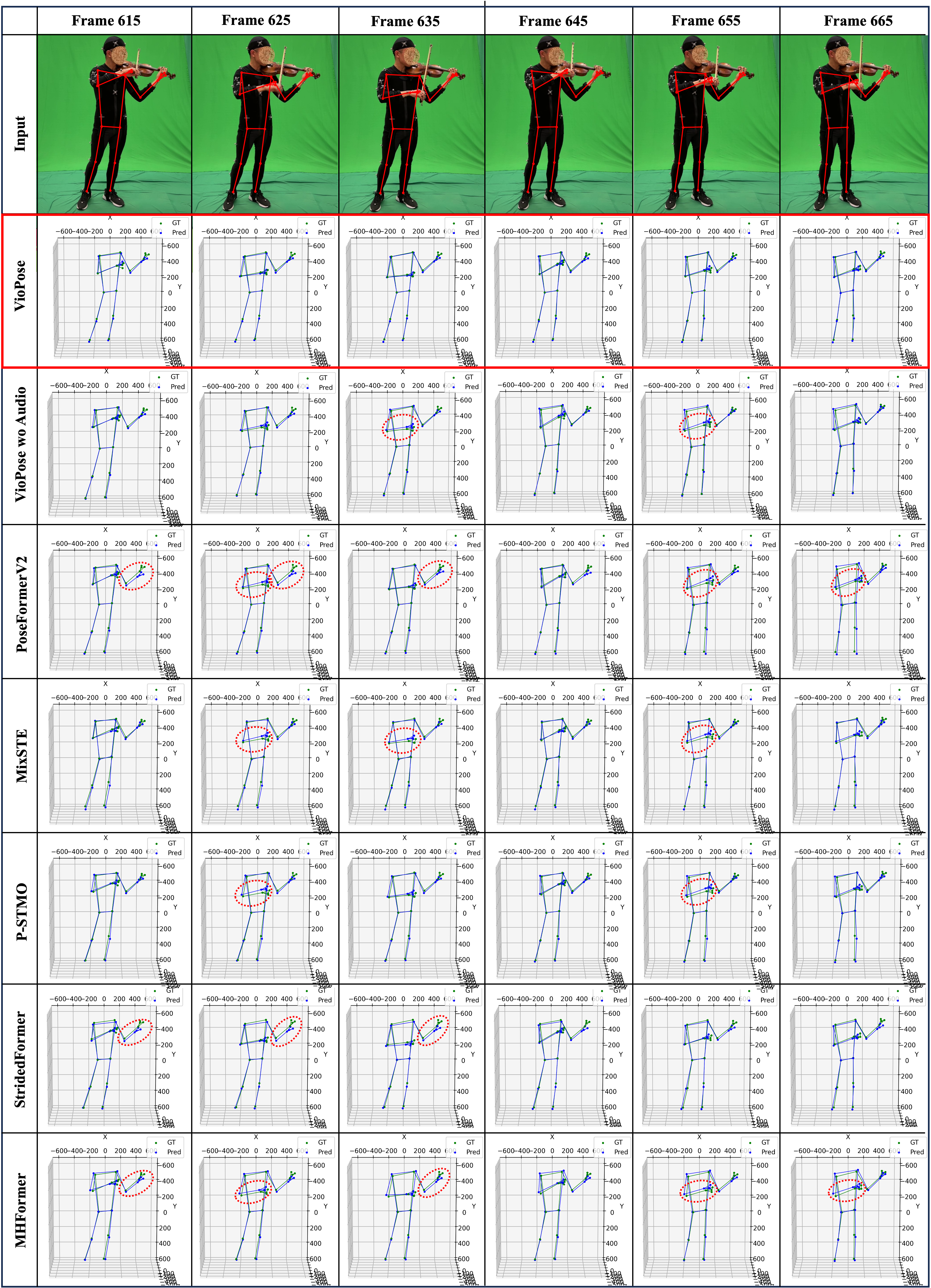}
	\caption{
	    Qualitative results, tested on \DataName \ P08C01T28 with 10 frames interval (30 FPS). Green and blue solid keypoints and are ground truth 3D pose and prediction, respectively. Red circles highlight incorrect estimations.
	}
	\label{fig:qualitative_rsult}
    \vspace{-4mm}
\end{figure}

We also illustrate the right and left wrist trajectories in Fig. ~\ref{fig:trajectory} and Fig.~\ref{fig:vibrato}, respectively. Furthermore,  we demonstrate a qualitative full-body estimation result in Fig.~\ref{fig:qualitative_rsult}. In Fig.~\ref{fig:trajectory}, we highlight the articulation of the estimated right-hand wrist trajectories compared to other SoTA methods. For example, most other SoTA methods decently estimate large and straightforward motions (first column of the figure) but tend to fail to estimate small (second column of the figure) and complex motions (third and last column of the figure). This is mainly because they achieve good MPJPE value even though they ignore fine grained motions. However, \AlgName \ explicitly learns motion dynamics via the hierarchical structure ($\S$ ~\ref{sec:hierarchy}) and bidirectional mixing module ($\S$ ~\ref{sec:mixing}), guiding the model to track sophisticated motions as well.
These phenomena are well observed for the case of vibrato in Fig. ~\ref{fig:vibrato}. Unlike bowing motion, vibrato lasts only a few seconds and has very fast and tiny motions (approximately 10 mm perturbation). Therefore, other SoTA methods tend to estimate simple straight lines rather than subtle perturbations. 

Note that, even though other algorithms estimate straight lines (or average trajectory), they still achieve approximately 10 mm error, which is less than the average MPJPE value. However, to gain a better understanding of human motion in musical performance, it is crucial to estimate those sophisticated motion dynamics as well. Especially in such cases, audio plays a significant role in compensating information, compared to \AlgName \ and \AlgName \ w/o Audio in Fig.~\ref{fig:vibrato}.
\begin{table}[t]
    \centering
    \resizebox{\columnwidth}{!}{
    \begin{tabular}{l | c|c|c|c|c|c}
        \Xhline{1pt}
        \textbf{Method} & \textbf{CNN} & \textbf{MPJPE} $\downarrow$ & \textbf{P-MPJPE} $\downarrow$  & \textbf{MPJVE} $\downarrow$  & \textbf{MPJAE} $\downarrow$  & \textbf{DTW} $\downarrow$ \\
        \hline
        Spectrogram (100 SR)
        & 1D & 44.01 & 28.01  & 1.56 & 1.03 &  6.70 
        \\
        30 SR
        & 1D & 44.11  & \textbf{27.39} & \textbf{1.55} & \textbf{1.01} &  6.74 
        \\
        100 SR (Baseline) 
        & 1D & 43.60 &  27.77 & 1.57 & 1.02 & \textbf{6.67 }
        \\
        100 SR
        & 2D & 45.01 & 27.98  & 1.64 & 1.08 & 6.83  
        \\
        100 SR
        & w/o & 45.39 & 28.69 & 1.62  & 1.08  & 6.84
        \\
        300 SR
        & 1D & \textbf{43.51}  & 27.86 & 1.65 & 1.10 & 6.69  
        \\
        \Xhline{1pt}
    \end{tabular}
    }
    \caption{Ablation study of different audio inputs: spectrogram (100 SR), 30, 100, 300 sampling rate audio features, 1D CNN, 2D CNN, and without CNN.}
    \label{tab:abl_audio}
    \vspace{-4mm}
\end{table}

\subsection{Ablation Study}
\label{sec:abl}
We conducted four ablation studies; specifically, we evaluated the audio module, the hierarchy module, the bidirectional mixing module, and different input frame sizes.

\paragraph{Audio Module}
We explored different types of audio inputs, sampling rates, and CNN models, as shown in Table ~\ref{tab:abl_audio}. We experimented with sampling rates ranging from 30 to 300 to assess their impacts. We also tested spectrogram (100 SR) inputs, standard in the audiovisual learning literature~\cite{suPhysicsDrivenDiffusionModels2023, suAudeoAudioGeneration2020}. In terms of metrics, 1D CNNs demonstrated superior results over 2D CNNs and basic MLPs (see \textit{w/o CNN} experiment in Table~\ref{tab:abl_audio}) for computing latent vector spaces. Increasing the sampling rate, the model showed better results in MPJPE. In conclusion, the key to performance improvement is using the mixture of audio features~\cite{liAIChoreographerMusic2021, marchellusM2CConciseMusic2023} with a 1D CNN feature extractor and a higher sampling rate. 

\begin{table}[t]
\vspace{-4mm}
    \centering
    \resizebox{\columnwidth}{!}{
    \begin{tabular}{l| c| c| c | c| c }
     \Xhline{1pt}
     \textbf{Method} & \textbf{MPJPE $\downarrow$} & \textbf{P-MPJPE $\downarrow$} & \textbf{MPJVE $\downarrow$} & \textbf{MPJAE $\downarrow$} & \textbf{DTW $\downarrow$} \\
    \hline
    \AlgName \  (Baseline)
    & \textbf{43.60} & \textbf{27.77} & 1.57  & \textbf{1.02} & \textbf{6.67}
    \\
    \hline
    w/o Cascade
    & 44.21 & 28.15 & 1.62 & 1.07 & 6.73
    \\
    Concatenation
    & 43.95 & 27.95 & \textbf{1.56} & \textbf{1.02} & 6.72
    \\
    Conditioning
    & 44.18 & 29.01 & 1.90 & 1.42 & 6.75 \\
    Parallel
    & 45.17 & 28.40 & 2.19 & 2.42 & 6.81 
    \\
    \hline
    w/o Int.
    & 44.75 & 28.44 & \textbf{1.57} & 1.04 & 6.78 \\ 
    w/o Diff.
    & 46.02 & 27.99 & 1.90 & 1.50 & 6.92 \\ 
    w/o Mixing
    & 47.87 & 29.61 & 2.00 & 1.34 & 7.10
    \\
    
    \Xhline{1pt}
    \end{tabular}
    }
    \caption{Ablation study on the efficacy of the modules in \AlgName. Bold numbers represent the best values.}
    \label{tab:abl}
\vspace{-4mm}
\end{table}
\paragraph{Hierarchy Module}
\label{sec:abl2}
We evaluated the efficacy of the hierarchy module (detailed in $\S$~\ref{sec:hierarchy}), as shown in Table ~\ref{tab:abl}. Our ablation study includes three scenarios. First, we examined a model without Eq.~\eqref{eq:sum}, referred to as \textit{w/o Cascade} in the table. 
Second, we replaced Eq.~\eqref{eq:sum} with a concatenation operator ($\|$):
\begin{align}
h_{i}^{(n)}= \text{Norm}(g_{i}^{(n)}\| h_{i+1}^{(n)}) 
\end{align}
Lastly, we removed Eq.~\eqref{eq:sum} and replaced Eq.~\eqref{eq:hierarchy} with a cross-transformer. Formally, 
\begin{align}
h_{i}^{(n)}=\text{Norm}(\text{CrossTransformer}(h_{i}^{(n-1)}, h_{i+1}^{(n)}))    
\end{align}
where $\text{CrossTransformer(x,y)}$ follows a computation similar as in Eq.~\eqref{eq:transformer}, but in the multi-head attention computation, $x$ becomes query and $y$ becomes key and value~\cite{attention}. 

As shown in Table~\ref{tab:abl}, the \textit{w/o Cascade} model shows the worst result in MPJPE. However, overall the performance drops are minor. This is mainly because the architecture maintains the hierarchy structure and explicitly exchanges information through the bidirectional mixing module. Once we remove the top-down integrated estimation (Eq. ~\eqref{eq:vel_int} and Eq.~\eqref{eq:pose_int}) and bottom-up differentiated estimation (Eq.~\eqref{eq:vel_diff} and Eq.~\eqref{eq:acc_diff}), the architecture becomes parallel and the performance drops significantly, especially in MPJVE and MPJAE values (\textit{Parallel} in Table~\ref{tab:abl}).

\begin{table}[b]
\vspace{-2mm}
    \centering
    \resizebox{\columnwidth}{!}{
    \begin{tabular}{l| c| c| c | c| c }
     \Xhline{1pt}
     \textbf{Input Frame} & \textbf{MPJPE $\downarrow$} & \textbf{P-MPJPE $\downarrow$} & \textbf{MPJVE $\downarrow$} & \textbf{MPJAE $\downarrow$} & \textbf{DTW $\downarrow$} \\
    \hline
    90
    & 44.22 & 29.02 & 1.73 & 1.43 & 6.73
    \\
    120
    & 44.60 & 27.79 & 1.62 & 1.24 & 6.77
    \\
    150  (Baseline)
    & \textbf{43.60} & \textbf{27.77} & \textbf{1.57}  & \textbf{1.02} & \textbf{6.67}
    \\
    180
    & 44.85 & 27.84 &  1.63 & 1.23 & 6.82
    \\
    
    \Xhline{1pt}
    \end{tabular}
    }
    \caption{Ablation study on the different input frame size. Bold numbers represent the best values.}
    \label{tab:abl_frame}
\end{table}
\paragraph{Bidirectional Mixing Module}
\label{sec:abl3}
We evaluated the efficacy of the bidirectional mixing module (\S ~\ref{sec:mixing}) as shown in Table ~\ref{tab:abl}. First, we removed Eq.~\eqref{eq:vel_int} and Eq.~\eqref{eq:pose_int}, referred to as \textit{w/o Int.}. Second, we removed Eq.~\eqref{eq:vel_diff} and Eq.~\eqref{eq:acc_diff}, referred to as \textit{w/o Diff.}. Lastly, we tested the model without the bidirectional mixing module. Based on the results, bottom-up motion dynamics mixing is important. We assume this is because the architecture already exchanges top-down information through the hierarchy module. However, explicit top-down motion dynamics mixing is still crucial to achieve better results. 

\paragraph{Different Input Frame Size}
Lastly, we changed the input frame size from 90 (3 sec) to 180 (6 sec), as shown in Table~\ref{tab:abl_frame}. Interestingly, the metrics show a U-shaped performance gain with more extended frames except in the MPJPE value. Unlike other architectures, we estimate the full frame output, which dilutes the importance of single-frame accuracy with longer input frame sequences. Therefore, we chose 150 frames as input.

\begin{table}[t]
    \centering
    \resizebox{\columnwidth}{!}{
    \begin{tabular}{l|c|c|c| c }
         \Xhline{1pt}
         \textbf{Method}& \textbf{Bow Dir (\%) $\uparrow$} & \textbf{Straight Bow (\%) $\uparrow$} & \textbf{Violin Hold (\%) $\uparrow$} & \textbf{Vibrato (\%)$\uparrow$}  \\
         \hline
             
        PoseFormerV2 (T=81)~\cite{zhaoPoseFormerV2ExploringFrequency2023} 
        & 55.04 & 7.16 & 24.32 & 52.03
        \\
        
        MixSTE (T=243) ~\cite{zhangMixSTESeq2seqMixed2022}
        & 53.94 & 2.13 & 30.24 & \textcolor{blue}{56.84}
        \\
        
        P-STMO (T=243) ~\cite{shanPSTMOPretrainedSpatial2022} 
        & 69.12 & \textcolor{blue}{35.15} & 14.97 & 52.03
        \\
        
        StridedFormer (T=351) ~\cite{liExploitingTemporalContexts2022}
        & 69.25 & 17.40 & 28.69 & 52.03
        \\
        
        MHFormer (T=351) ~\cite{liMHFormerMultiHypothesisTransformer2022} 
        & 49.22 & 2.09 & \textcolor{blue}{34.03} & 51.50
        \\
        \hline
        \AlgName \ w/o audio (T=150) 
        & \textcolor{blue}{69.26} & 31.75 & 33.36 & 52.22
        \\
        \AlgName \  (T=150)
        & \textcolor{red}{71.49} & \textcolor{red}{41.51} & \textcolor{red}{35.46} & \textcolor{red}{63.33}
        \\
        
         \Xhline{1pt}
    \end{tabular}
    }
    \caption{Violin performance analysis results. We computed the mean value and standard deviation over trials. \textcolor{red}{Red} and \textcolor{blue}{blue} represent first and second best, respectively.}
    \label{tab:result_performance_wild}
    \vspace{-4mm}
\end{table}
\subsection{Violin Performance Analysis}
\label{sec:violin_performance_analysis}
\AlgName\ will be used in an AI system giving feedback to violin players. We evaluated four tasks to showcase the algorithm's usefulness for violin performance analysis. First, we used it for segmenting bowing direction change, essential for other downstream tasks related to bow stroke analysis. We low-pass filtered the right wrist estimates and computed the local minima of the acceleration to locate the bowing direction change. Comparison with ground truth is within a temporal window of acceptance. 
Second, we monitored the bowing trajectory. Ideally, the bow is parallel to the bridge (we call this a straight bow in Table \ref{tab:result_performance_wild}). For each segmented bow stroke, we measured the correctness of the curvature of the right wrist trajectory.
Thirdly, we monitored the left wrist flexion (useful for evaluating violin hold) from the hand, wrist, and elbow positions and evaluated using the $L_1$ distance to the ground truth. Fourthly, we computed vibrato.
We computed local minima of acceleration of the mid-point of the left wrist, pinkie, and index finger, to obtain location direction change and computed temporal alignment with the ground truth.

\AlgName\ outperforms the SoTA in all four tasks as shown in Table \ref{tab:result_performance_wild}. For segmenting bowing direction change, \AlgName\ is 3.1\% to 31.2\% more accurate than the SoTA and 3.1\% better than without audio. For the straight bow analysis, \AlgName\  was 15.3\% to 95.0\% more accurate than the SoTA and 23.5\% more accurate than without the audio model. For violin hold, \AlgName \ is 4.0\% to 57.8\% more accurate and 5.9\% improvement with audio. And finally, for vibrato, \AlgName \  is 10.2\% to 17.8\% more accurate than SoTA and 17.5\% improvement over the non-audio model. These results clearly demonstrate why a more accurate and stable 3D pose estimation is needed and across the board audio contributes to much better model prediction.


\section{Conclusion}
\label{sec:conclusion}
We introduced a large violin dataset, captured with four cameras and paired microphones and synchronized and calibrated with the 2D and 3D poses. It features players of diverse genders, ages, sizes, and expertise who play common practice materials and complex repertoires. We demonstrated the shortfalls of existing SoTA pose estimation algorithms for violin playing. We then presented an audiovisual monocular 4D pose estimation network with a novel hierarchical structure outperforming current SoTA methods, and demonstrated the usefulness of the method for violin performance analysis. In the next step in our future work, we plan to analyze and segment the video into elementary segments \cite{guerra2005discovering,li2010learning}, which are meaningful to music performance.

\section{Acknowledgements}
We greatly acknowledge NSF's support under awards OISE 2020624 and BCS 2318255 and UMD's Grand Challenge Grant for "Music Education for All Through Personalized AI and Digital Humanities".

{\small
\bibliographystyle{IEEEtranSN}
\bibliography{egbib}
}

\end{document}